\documentclass{article}

\PassOptionsToPackage{numbers, square}{natbib}

\usepackage[preprint]{nips_2018}

\usepackage[utf8]{inputenc} 
\usepackage[T1]{fontenc}    
\usepackage{hyperref}       
\usepackage{url}            
\usepackage{booktabs}       
\usepackage{amsmath,amsfonts,amssymb}       
\usepackage{nicefrac}       
\usepackage{microtype}      
\usepackage{csquotes}
\usepackage{array}

\usepackage{wrapfig}
\usepackage{graphbox}
\usepackage{subcaption}
\usepackage{tikz}
\usetikzlibrary{arrows}

\graphicspath{{./}{./figures/}}

\newcommand{\act}{a}
\newcommand{\preact}{h}
\newcommand{\err}{E}
\DeclareMathOperator{\diag}{diag}

\newcommand{\data}{\mathcal{D}}
\newcommand{\weight}{W}
\newcommand{\weights}{\theta}
\DeclareMathOperator*{\argmax}{arg\,max}
\newcommand{\hess}{H}

\newcommand{\deriv}[2]{\frac{\partial #1}{\partial #2}}
\newcommand{\dderiv}[3]{\frac{\partial^2 #1}{\partial #2 \partial #3}}
\newcommand{\ederiv}[1]{\deriv{\err}{#1}}
\newcommand{\ehess}[2]{\dderiv{\err}{#1}{#2}}

\newcommand{\evalat}[2]{\left.#1\right\rvert_{#2}}

\newcommand{\actf}{\mathcal{Q}}
\newcommand{\xhess}{\mathcal{H}}
\DeclareMathOperator{\vect}{vec}

\newcommand{\fref}[1]{Fig.~\ref{#1}}

\newcommand{\sref}[1]{Section~\ref{#1}}
\newcommand{\aref}[1]{Appendix~\ref{#1}}
\newcommand{\eref}[1]{Eq.~(\ref{#1})}

\title{Online Structured Laplace Approximations For Overcoming Catastrophic Forgetting}

\author{
  Hippolyt Ritter\thanks{Corresponding author: \texttt{j.ritter@cs.ucl.ac.uk}} \\
  University College London
  \And
  Aleksandar Botev\\
  University College London
  \And
  David Barber\\
  University College London \\\& Alan Turing Institute
}

\begin{document}

\maketitle

\begin{abstract}
We introduce the Kronecker factored online Laplace approximation for overcoming catastrophic forgetting in neural networks.
The method is grounded in a Bayesian online learning framework, where we recursively approximate the posterior after every task with a Gaussian, leading to a quadratic penalty on changes to the weights.
The Laplace approximation requires calculating the Hessian around a mode, which is typically intractable for modern architectures.
In order to make our method scalable, we leverage recent block-diagonal Kronecker factored approximations to the curvature.
Our algorithm achieves over $90\%$ test accuracy across a sequence of $50$ instantiations of the permuted MNIST dataset, substantially outperforming related methods for overcoming catastrophic forgetting.
\end{abstract}

\section{Introduction} \label{sec:intro}

Creating an agent that performs well across multiple tasks and continuously incorporates new knowledge has been a longstanding goal of research on artificial intelligence.
When training on a sequence of tasks, however, the performance of many machine learning algorithms, including neural networks, decreases on older tasks when learning new ones.
This phenomenon has been termed `catastrophic forgetting' \citep{catastrophic:forgetting:connectionist,catastrophic:interference,connectionist:models:memory} and has recently received attention in the context of deep learning \citep{empirical:cf,ewc}.
Catastrophic forgetting cannot be overcome by simply initializing the parameters for a new task with optimal ones from the old task and hoping that stochastic gradient descent will stay sufficiently close to the original values to maintain good performance on previous datasets \citep{empirical:cf}.

Bayesian learning provides an elegant solution to this problem.
It combines the current data with prior information to find an optimal trade-off in our belief about the parameters.
In the sequential setting, such information is readily available: the posterior over the parameters given all previous datasets.
It follows from Bayes' rule that we can use the posterior over the parameters after training on one task as our prior for the next one.
As the posterior over the weights of a neural network is typically intractable, we need to approximate it.
This type of Bayesian online learning has been studied extensively in the literature \citep{bayesian:online,online:variational:bayes:zoubin,online:variational:bayes:honkela}.

In this work, we combine Bayesian online learning \citep{bayesian:online} with the Kronecker factored Laplace approximation \citep{scalable:laplace} to update a quadratic penalty for every new task.
The block-diagonal Kronecker factored approximation of the Hessian \citep{kfac,kfra} allows for an expressive scalable posterior that takes interactions between weights within the same layer into account.
In our experiments we show that this principled approximation of the posterior leads to substantial gains in performance over simpler diagonal methods, in particular for long sequences of tasks.

\section{Bayesian online learning for neural networks} \label{sec:ewc}

We are interested in optimizing the parameters $\theta$ of a single neural network to perform well across multiple tasks $\data_1, \ldots, \data_T$, specifically finding a MAP estimate $\weights^* = \argmax_\weights p(\weights|\data_1, \ldots, \data_T)$.
However, the datasets arrive sequentially and we can only train on one of them at a time.

In the following, we first discuss how Bayesian online learning solves this problem and introduce an approximate procedure for neural networks.
We then review recent Kronecker factored approximations to the curvature of neural networks and how to use them to obtain a better fit to the posterior.
Finally, we introduce a hyperparameter that acts as a regularizer on the approximation to the posterior.

\subsection{Bayesian online learning}

Bayesian online learning \citep{bayesian:online}, or Assumed Density Filtering \citep{smec}, is a framework for updating an approximate posterior when data arrive sequentially.
Using Bayes' rule we would like to simply incorporate the most recent dataset $\data_{t+1}$ into the posterior as:

\begin{equation}
    p(\weights|\data_{1:t+1})
    = \frac{p(\data_{t+1}|\weights) p(\weights|\data_{1:t})}
           {\int d\weights' p(\data_{t+1}|\weights') p(\weights'|\data_{1:t})}
\end{equation}

where we use the posterior $p(\weights|\data_{1:t})$ from the previously observed tasks as the prior over the parameters for the most recent task.
As the posterior given the previous datasets is typically intractable, Bayesian online learning formulates a parametric approximate posterior $q$ with parameters $\phi_t$, which it iteratively updates in two steps:

\paragraph{Update step}

In the update step, the approximate posterior $q$ with parameters $\phi_t$ from the previous task is used as a prior to find the new posterior given the most recent data:

\begin{equation}
    p(\weights|\data_{t+1}, \phi_t)
    = \frac{p(\data_{t+1}|\weights) q(\weights|\phi_t)}
           {\int d\weights' p(\data_{t+1}|\weights') q(\weights'|\phi_t)}
    \label{eq:bo:update}
\end{equation}

\paragraph{Projection step}

The projection step finds the distribution within the parametric family of the approximation that most closely resembles this posterior, i.e. sets $\phi_{t+1}$ such that:

\begin{equation}
    q(\weights|\phi_{t+1}) \approx p(\weights|\data_{t+1}, \phi_t)
\end{equation}

\citet{bayesian:online} suggest minimizing the KL-divergence between the approximate and the true posterior, however this is mostly appropriate for models where the update-step posterior and a solution to the KL-divergence are available in closed form.
In the following, we therefore propose using a Laplace approximation to make Bayesian online learning tractable for neural networks.

\subsection{The online Laplace approximation}

Neural networks have found wide-spread success and adoption by performing simple MAP inference, i.e. finding a mode of the posterior:

\begin{equation}
        \weights^*
        = \argmax_\weights \; \log p(\weights|\data)
        = \argmax_\theta \; \log p(\data|\weights) + \log p(\weights)
        \label{eq:post}
\end{equation}

where $p(\data|\weights)$ is the likelihood of the data and $p(\weights)$ the prior.
Most commonly used loss functions and regularizers fit into this framework, e.g. using a categorical cross-entropy with $L_2$-regularization corresponds to modeling the data with a categorical distribution and placing a zero-mean Gaussian prior on the network parameters.
A local mode of this objective function can easily be found using standard gradient-based optimizers.

Around a mode, the posterior can be locally approximated using a second-order Taylor expansion, resulting in a Normal distribution with the MAP parameters as the mean and the Hessian of the negative log posterior around them as the precision.
Using a Laplace approximation for neural networks was pioneered by \citet{mackay:laplace}.

We therefore proceed in two iterative steps similar to Bayesian online learning, using a Gaussian approximate posterior for $q$, such that $\phi_t = \{\mu_t, \Lambda_t\}$ consists of a mean $\mu$ and a precision matrix $\Lambda$:

\paragraph{Update step}

As the posterior of a neural network is intractable for all but the simplest architectures, we will work with the unnormalized posterior.
The normalization constant is not needed for finding a mode or calculating the Hessian.
The Gaussian approximate posterior results in a quadratic penalty encouraging the parameters to stay close to the mean of the previous approximate posterior:

\begin{equation}
    \begin{split}
        \log p(\weights|\data_{t+1}, \phi_t)
        &\propto \log p(\data_{t+1}|\weights) + \log q(\weights|\phi_t)\\
        &\propto \log p(\data_{t+1}|\weights) - \frac{1}{2} (\weights - \mu_t)^\top \Lambda_t (\weights - \mu_t)
        \label{eq:objective}
    \end{split}
\end{equation}


\paragraph{Projection step}

In the projection step we approximate the posterior with a Gaussian.
We first update the mean of the approximation to a mode of the new posterior:

\begin{equation}
    \mu_{t+1} = \argmax_\weights \; \log p(\data_{t+1}|\weights) + \log q(\weights|\phi_t)
\end{equation}

and then perform a quadratic approximation around it, which requires calculating the Hessian of the negative objective.
This leads to a recursive update to the precision with the Hessian of the most recent log likelihood, as the Hessian of the negative log approximate posterior is its precision:

\begin{equation}
    \Lambda_{t+1} = \hess_{t+1}(\mu_{t+1}) + \Lambda_t
\end{equation}

where $\hess_{t+1}(\mu_{t+1}) = -\evalat{\dderiv{\log p(\data_{t+1}|\weights)}{\weights}{\weights}}{\weights=\mu_{t+1}}$ is the Hessian of the newest negative log likelihood around the mode.
The precision of a Gaussian is required to be positive semi-definite, which is the case for the Hessian at a mode.
In order to numerically guarantee this in practice, we use the Fisher Information as an approximation \citep{fisher} that is positive semi-definite by construction.

The recursion is initialized with the Hessian of the log prior, which is typically constant.
For a zero-mean isotropic Gaussian prior, corresponding to an $L_2$-regularizer, it is simply the identity matrix times the prior precision.\footnote{\citet{quadratic:penalties} recently discussed a similar recursive Laplace approximation for online learning, however with limited experimental results and in the context of using a diagonal approximation to the Hessian.}

A desirable property of the Laplace approximation is that the approximate posterior becomes peaked around its current mode as we observe more data.
This becomes particularly clear if we think of the precision matrix as the product of the number of data points and the average precision.
By becoming increasingly peaked, the approximate posterior will naturally allow the parameters to change less for later tasks.
At the same time, even though the Laplace method is a local approximation, we would expect it to leave sufficient flexibility for the parameters to adapt to new tasks, as the Hessian of neural networks has been observed to be flat in most directions \citep{hessian:empirical}.

We will also compare to fitting the true posterior with a new Gaussian at every task for which we compute the Hessian of all tasks around the most recent MAP estimate:

\begin{equation}
    \Lambda_{t+1} = \hess_{prior} + \sum_{i=1}^{t+1} \hess_i(\mu_{t+1})
    \label{eq:approximate}
\end{equation}

This procedure differs from the online Laplace approximation only in evaluating all Hessians at the most recent MAP parameters instead of the respective task's ones.
Technically, this is not a valid Laplace approximation, as we only optimize an approximation to the posterior.
Hence the optimal parameters for the approximate objective will not exactly correspond to a mode of the exact posterior. 
However, as we will use a positive semi-definite approximation to the Hessian, this will only introduce a small additional approximation error.

Calculating the Hessian across all datasets requires relaxing the sequential learning setting to allowing access to previous data `offline', i.e. between tasks.
We use this baseline to check if there is any loss of information in using estimates of the curvature at previous parameter values.

\subsection{Kronecker factored approximation of the Hessian}

Modern networks typically have millions of parameters, so the size of the Hessian is several terabytes.
An approximation that is simple to implement with automatic differentiation frameworks is the diagonal of the Fisher matrix, i.e. the expected square of the gradients, where the expectation is over the datapoints and the conditional distribution defined by the model.
While this approximation has been used successfully \citep{ewc}, it ignores interactions between the parameters.

Recent works on second-order optimization \citep{kfac,kfra} have developed block-diagonal approximations to the Hessian.
They exploit that, for a single data point, the diagonal blocks of the Hessian of a feedforward network --- corresponding to the weights of a single layer --- are Kronecker factored, i.e. a product of two relatively small matrices.

We denote a neural network as taking an input $\act_0{=}x$ and producing an output $\preact_L$.
The input is passed through layers $1,\ldots,L$ as the linear pre-activations $\preact_l{=}\weight_l \act_{l-1}$ and the activations $\act_l{=}f_l(\preact_l)$, where $f_l$ is a non-linear elementwise function.
The outputs then parameterize the log likelihood of the data, and, using the chain rule, we can write the Hessian w.r.t. the weights of a single layer as:

\begin{equation}
    \hess_l = \dderiv{\log p(\data|\preact_L)}{\vect(\weight_l)}{\vect(\weight_l)} = \actf_l \otimes \xhess_l \label{eq:kf}
\end{equation}

where $\vect(\weight_l)$ is the weight matrix of layer $l$ stacked into a vector and we define $\actf_l = a_{l-1} a_{l-1}^\top$ as the covariance of the inputs to the layer.
$\xhess_l = \dderiv{\log p(\data|\weights)}{h_l}{h_l}$ is the pre-activation Hessian, i.e. the second derivative w.r.t. the pre-activations $h_l$ of the layer.
We provide the basic derivation of \eref{eq:kf} and the recursive formula for calculating $\xhess_l$ in \aref{app:kf}.
To maintain the Kronecker factorization in expectation, i.e. for an entire dataset, \citep{kfac} and \citep{kfra} assume the two factors to be independent and approximate the expected Kronecker product by the Kronecker product of the expected factors.

The block-diagonal approximation splits the Hessian-vector product in the quadratic penalty across the layers.
Due to the Kronecker factored approximation, it can be calculated efficiently for each layer using the following well-known identity:

\begin{equation}
    (\actf_l \otimes \xhess_l) \vect(\weight_l - \weight_l^*)
    = \vect(\xhess_l \left(\weight_l - \weight_l^*\right)^\top \actf_l)
\end{equation}

where $\vect$ stacks the columns of a matrix into a vector and we use that $\xhess$ is symmetric.

The block-diagonal Kronecker factored approximation corresponds to assuming independence between the layers and factorizing the covariance between the weights of a layer into the covariance of the columns and rows, resulting in a matrix normal distribution \citep{matrixnormal}.
The same approximation has been used recently to sample from the predictive posterior \citep{scalable:laplace,bayesian:maml}.
While it still makes some independence assumptions about the weights, the most important interactions --- the ones within the same layer --- are accounted for.
In order to guarantee for the curvature being positive semi-definite, we approximate the Hessian with the Fisher Information as in \citep{kfac} throughout our experiments.

\subsection{Regularizing the approximate posterior} \label{sec:hyperparam}

\citet{ewc}, who develop a similar method inspired by the Laplace approximation, suggest using a multiplier $\lambda$ on the quadratic penalty in \eref{eq:objective}.
This hyperparameter provides a way of trading off retaining performance on previous tasks against having sufficient flexibility for learning a new one.
As modifying the objective would propagate into the recursion for the precision matrix, we instead place the multiplier on the Hessian of each log likelihood and update the precision as:

\begin{equation}
    \Lambda_{t+1} = \lambda \hess_{t+1}(\mu_{t+1}) + \Lambda_t
\end{equation}

The multiplier affects the width of the approximate posterior and thus the location of the next MAP estimate.
As it acts directly on the parameter of a probability distribution, its optimal value can inform us about the quality of our approximation: if it strongly deviates from its natural value of $1$, our approximation is a poor one and over- or underestimates the uncertainty about the parameters.
We visualize the effect of $\lambda$ in \fref{fig:contours} in \aref{app:reg:vis}.

\section{Related work} \label{sec:related}

Our method is closely related to Bayesian online learning \citep{bayesian:online} and to Laplace propagation \citep{laplace:prop}.
In contrast to Bayesian online learning, as we cannot update the posterior over the weights in closed form, we use gradient-based methods to find a mode and perform a quadratic approximation around it, resulting in a Gaussian approximation.
Laplace propagation, similar to expectation propagation \citep{ep}, maintains a factor for every task, but approximates each of them with a Gaussian.
It performs multiple updates, whereas we use each dataset only once to update the approximation to the posterior.

The most similar method to ours for overcoming catastrophic forgetting is Elastic Weight Consolidation (EWC) \citep{ewc}.
EWC approximates the posterior after the first task with a Gaussian.
However, it continues to add a penalty for every new task \citep{reply:ferenc}.
This is more closely related to Laplace propagation, but may be overcounting early tasks \citep{quadratic:penalties} and does not approximate the posterior.
Furthermore, EWC uses a simple diagonal approximation to the Hessian.
\citet{imm} approximate the posterior around the mode for each dataset with a diagonal Gaussian in addition to a similar approximation of the overall posterior.
They update this approximation to the posterior as the Gaussian that minimizes the KL divergence with the individual posterior approximations.
\citet{vcl} implement online variational learning \citep{online:variational:bayes:zoubin,online:variational:bayes:honkela}, which fits an approximation to the posterior through the variational lower bound and then uses this approximation as the prior on the next task.
Their Gaussian is fully factorized, hence they do not take weight interactions into account either.

\citep{scalable:laplace} and \citep{bayesian:maml} have independently proposed the use of block-diagonal Kronecker factored curvature approximations \citep{kfac,kfra} to sample from an approximate Gaussian posterior over the weights of a neural network.
They find that this requires adding a multiple of the identity to their curvature factors as an ad-hoc regularizer, which is not necessary for our method.
In our work, we use an approximate posterior with the same Kronecker factored covariance structure as a prior for subsequent tasks.
We iteratively update this approximation for every new dataset.
The curvature factors that we accumulate throughout training could be used on top of our method to approximate the predictive posterior similar to \citep{scalable:laplace,bayesian:maml}.
However, both the curvature factors and the mode that our method finds will be different to performing a Laplace approximation in batch mode.
Our work links the Kronecker factored Laplace approximation \citep{scalable:laplace} to Bayesian online learning \citep{bayesian:online} similar to how Variational Continual Learning \citep{vcl} connects Online Variational Learning \citep{online:variational:bayes:zoubin,online:variational:bayes:honkela} to Bayes-by-Backprop \citep{weight:uncertainty}.

We discuss additional related methods without a Bayesian motivation in \aref{app:related}.

\section{Experiments} \label{sec:experiments}

In our experiments we compare our online Laplace approximation to the approximate Laplace approximation of \eref{eq:approximate} as well as EWC \citep{ewc} and Synaptic Intelligence (SI) \citep{synaptic:intelligence}, both of which also add quadratic regularizers to the objective.
Further, we investigate the effect of using a block-diagonal Kronecker factored approximation to the curvature over a diagonal one.
We also run EWC with a Kronecker factored approximation, even though the original method is based on a diagonal one.
We implement our experiments using Theano \citep{theano} and Lasagne \citep{lasagne} software libraries.

\subsection{Permuted MNIST}

\begin{figure}[ht!]
    \begin{minipage}{0.475\textwidth}
        \centering
        \includegraphics[width=\linewidth,align=t]{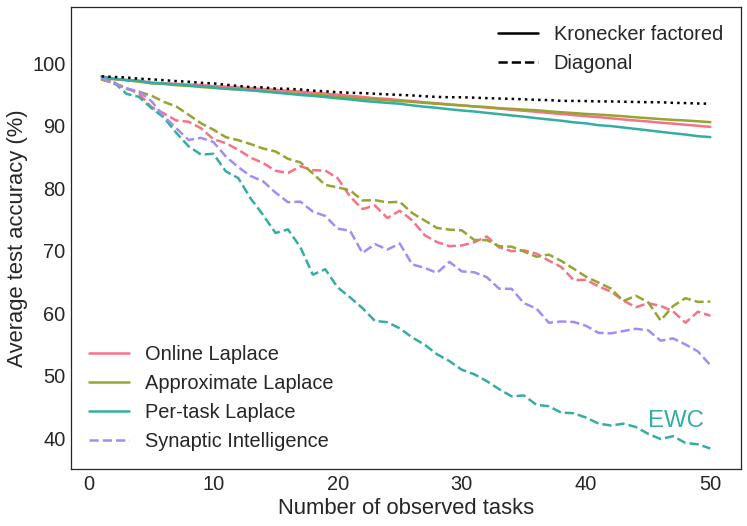}
        \caption{Mean test accuracy on a sequence of permuted MNIST datasets. We categorize SI as a diagonal method, as it does not account for parameter interactions. The dotted black line shows the performance of a single network trained on all observed data at each task.}
        \label{fig:pmnist:accs}
    \end{minipage}%
    \hfill%
    \begin{minipage}{0.475\textwidth}
        \centering
        \begin{subfigure}[]{\textwidth}
            \includegraphics[width=\textwidth,align=t]{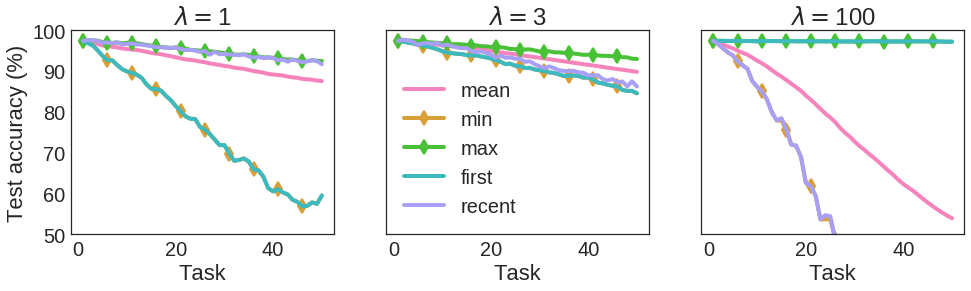}
            \caption{Kronecker factored}
            \label{fig:pmnist:hyper:kf}
        \end{subfigure}%
        \\%
        \begin{subfigure}[]{\textwidth}
            \includegraphics[width=\textwidth]{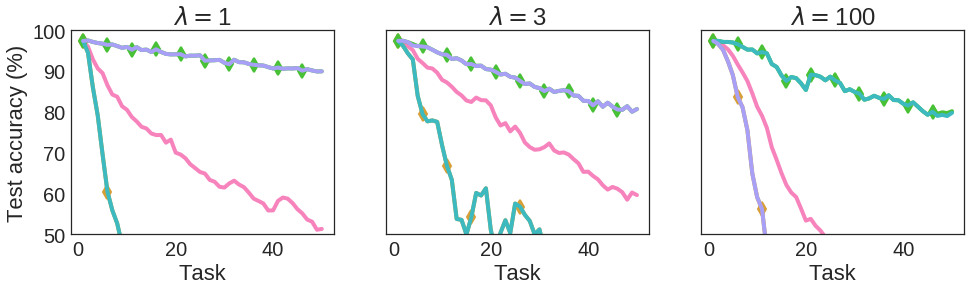}
            \caption{Diagonal}
            \label{fig:pmnist:hyper:diag}
        \end{subfigure}
        \caption{Effect of $\lambda$ for different curvature approximations for permuted MNIST. Each plot shows the mean, minimum and maximum across the tasks observed so far, as well as the accuracy on the first and most recent task.}
        \label{fig:pmnist:hyper}
    \end{minipage}
\end{figure}

As a first experiment, we test on a sequence of permutations of the MNIST dataset \citep{mnist}.
Each instantiation consists of the $28{\times}28$ grey-scale images and labels from the original dataset with a fixed random permutation of the pixels.
This makes the individual data distributions mostly independent of each other, testing the ability of each method to fully utilize the model's capacity.

We train a feed-forward network with two hidden layers of $100$ units and ReLU nonlinearities on a sequence of $50$ versions of permuted MNIST.
Every one of these datasets is equally difficult for a fully connected network due to its permutation invariance to the input.
We stress that our network is smaller than in previous works as the limited capacity of the network makes the task more challenging.
Further, we train on a longer sequence of datasets.
Optimization details are in \aref{app:details}.

\fref{fig:pmnist:accs} shows the mean test accuracy as new datasets are observed for the optimal hyperparameters of each method.
We refer to the online Laplace approximation as `Online Laplace', to the Laplace approximation around an approximate mode as `Approximate Laplace' and to adding a quadratic penalty for every set of MAP parameters as in \citep{ewc} as `Per-task Laplace'.
The per-task Laplace method with a diagonal approximation to the Hessian corresponds to EWC.

We find our online Laplace approximation to maintain higher test accuracy throughout training than placing a quadratic penalty around the MAP parameters of every task, in particular when using a simple diagonal approximation to the Hessian.
However, the main difference between the methods lies in using a Kronecker factored approximation of the curvature over a diagonal one.
Using this approximation, we achieve over $90\%$ average test accuracy across $50$ tasks, almost matching the performance of a network trained jointly on all observed data.
Recalculating the curvature for each task instead of retaining previous estimates does not significantly affect performance.

Beyond simple average performance, we investigate different values of the hyperparameter $\lambda$ on the permuted MNIST sequence of datasets for our online Laplace approximation.
The goal is to visualize how it affects the trade-off between remembering previous tasks and being able to learn new ones for the two approximations of the curvature that we consider.
\fref{fig:pmnist:hyper} shows various statistics of the accuracy on the test set for the smallest and largest value of the hyperparameter on the quadratic penalty that we tested, as well as the one that optimizes the validation error.

We are particularly interested in the performance on the first dataset and the most recent one, as a measure for memory and flexibility respectively.
For all displayed values of the hyperparameter, the Kronecker factored approximation (\fref{fig:pmnist:hyper:kf}) has higher test accuracy than the diagonal approximation (\fref{fig:pmnist:hyper:diag}) on both the most recent and the first task, as well as on average.
For the natural choice of $\lambda=1$ (leftmost subfigure respectively), the network's performance decays for the first task for both curvature approximations, yet it is able to learn the most recent task well.
The performance on the first task decays more slowly, however, for the more expressive Kronecker factored approximation of the curvature.
Increasing the hyperparameter, corresponding to making the prior more narrow as discussed in \sref{sec:hyperparam}, leads to the network remembering the first task much better at the cost of not being able to achieve optimal performance on the most recently added task.
Using  $\lambda=3$ (central subfigure), the value that achieves optimal validation error in our experiments, the Kronecker factored approximation leads to the network performing similarly on the most recent and first tasks.
This coincides with optimal average test accuracy.
We are not able to find such an ideal trade-off for the diagonal Hessian approximation, resulting in worse average performance and suggesting that the posterior cannot be matched well without accounting for interactions between the weights.
Using a large value of $\lambda=100$ (rightmost subfigure) reverts the order of performance between the most recent and the first task for both approximations: while for small $\lambda$ the first task is `forgotten', the network's performance now stays at a high level --- for the Kronecker factored approximation it remembers it perfectly --- which comes at the cost of being unable to learn new tasks well.

We conclude from our results that the online Laplace approximation overestimates the uncertainty in the approximate posterior about the parameters for the permuted MNIST task, in particular with a diagonal approximation to the Hessian.
Overestimating the uncertainty leads to a need for regularization in the form of reducing the width of the approximate posterior, as the value that optimizes the validation error is $\lambda=3$.
Only when regularizing too strongly the approximate posterior underestimates the uncertainty about the weights, leading to reduced performance on new tasks for large values of $\lambda$.
Using a better approximation to the posterior leads to a drastic increase in performance and a reduced need for regularization in the subsequent experiments.
We note that some regularization is still necessary, suggesting that even the Kronecker factored approximation overestimates the variance in the posterior, and a better approximation could lead to further improvements.
However, it is also possible that the Laplace approximation as such requires a large amount of data to estimate the interaction between the parameters sufficiently well; hence it might be best suited for settings where plenty of data are available.

\subsection{Disjoint MNIST}

\begin{wrapfigure}{r}{0.35\textwidth}
    \centering
    \includegraphics[width=\linewidth]{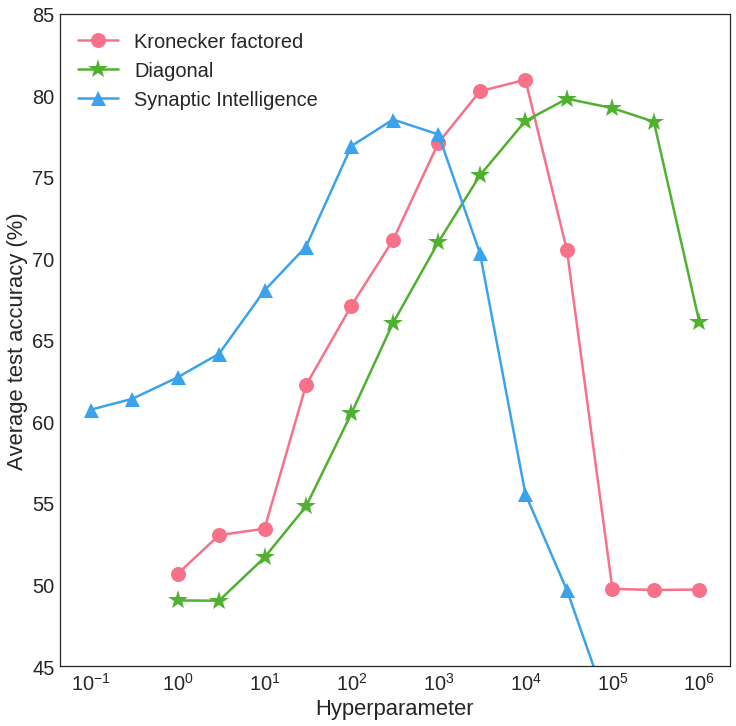}
    \caption{Disjoint MNIST test accuracy for the Laplace approximation (hyperparameter: $\lambda$) and SI (hyperparameter: $c$). `Kronecker factored' and `Diagonal' refer to the respective curvature approximation for the Laplace method.}
    \label{fig:disjoint}
\end{wrapfigure}

We further experiment with the disjoint MNIST task, which splits the MNIST dataset into one part containing the digits `$0$' to `$4$', and a second part containing `$5$' to `$9$' and training a ten-way classifier on each set separately.
Previous work \citep{imm} has found this problem to be challenging for EWC, as during the first half of training the network is encouraged to set the bias terms for the second set of labels to highly negative values.
This setup makes it difficult to balance out the biases for the two sets of classes after the first task without overcorrecting and setting the biases for the first set of classes to highly negative values.
\citet{imm} report just over $50\%$ test accuracy for EWC, which corresponds to either completely forgetting the first task or being unable to learn the second one, as each task individually can be solved with around $99\%$ accuracy.

We use an identical network architecture to the previous section and found stronger regularization of the approximate posterior to be necessary.
For the Laplace methods, we tested values of $\lambda \in \{1, 3, 10, \ldots, 3{\times}10^5,10^6\}$, and $c \in \{0.1, 0.3, 1, \ldots, 3{\times}10^4,10^5\}$ for SI.
We train using Nesterov momentum with a learning rate of $0.1$ and momentum of $0.9$ and decay the learning rate by a factor of $10$ every $1000$ parameter updates using a batch size of $250$.
We decay the initial learning rate for the second task depending on the hyperparameter to prevent the objective from diverging.
We test various decay factors for each hyperparameter, but as a rule of thumb found $\frac{\lambda}{10}$ to perform well for the Kronecker factored, and $\frac{\lambda}{1000}$ for the diagonal approximation.
The results are averaged across ten independent runs.

\fref{fig:disjoint} shows the test accuracy for various hyperparameter values for a Kronecker factored and a diagonal approximation of the curvature as well as SI.
As there are only two datasets, the three Laplace-based methods are identical, therefore we focus on the impact of the curvature approximation.
Approximating the Hessian with a diagonal corresponds to EWC.
While we do not match the performance of the method developed in \citep{imm}, we find the Laplace approximation to work significantly better than reported by the authors.
The Kronecker factored approximation gives a small improvement over the diagonal one and requires weaker regularization, which further suggests that it better fits the true posterior.
It also outperforms SI.

\subsection{Vision datasets}

\begin{figure*}[!ht]
    \centering
    \includegraphics[width=\textwidth]{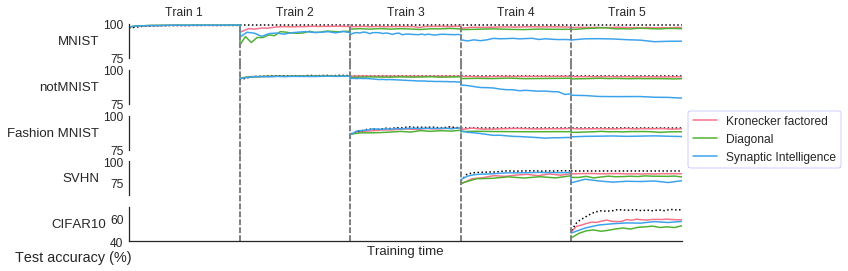}
    \caption{Test accuracy of a convolutional network on a sequence of vision datasets. We train on the datasets separately in the order displayed from top to bottom and show the network's accuracy on each dataset once training on it has started. The dotted black line indicates the performance of a network with the same architecture trained separately on the task. The diagonal and Kronecker factored approximation to the Hessian both use our online Laplace method to prevent forgetting.}
    \label{fig:lenet}
\end{figure*}

As a final experiment, we test our method on a suite of related vision datasets.
Specifically, we train and test on MNIST \citep{mnist}, notMNIST\footnote{Originally published at \url{http://yaroslavvb.blogspot.co.uk/2011/09/notmnist-dataset.html} and downloaded from \url{https://github.com/davidflanagan/notMNIST-to-MNIST}}, Fashion MNIST \citep{fashionmnist}, SVHN \citep{svhn} and CIFAR10 \citep{cifar} in this order.
All five datasets contain around $50,000$ training images from $10$ different classes.
MNIST contains hand-written digits from `$0$' to `$9$', notMNIST the letters `A' to `J' in different computer fonts, Fashion MNIST different categories of clothing, SVHN the digits `0' to `9' on street signs and CIFAR10 ten different categories of natural images.
We zero-pad the images of the MNIST-like datasets to be of size $32{\times}32$ and replicate their intensity values over three channels, such that all images have the same format.

We train a LeNet-like architecture \citep{mnist} with two convolutional layers with $5{\times}5$ convolutions with $20$ and $50$ channels respectively and a fully connected hidden layer with $500$ units.
We use ReLU nonlinearities and perform a $2{\times}2$ max-pooling operation after each convolutional layer with stride $2$.
An extension of the Kronecker factored curvature approximations to convolutional neural networks is presented in \citep{kfac:conv}. 
As the meaning of the classes in each dataset is different, we keep the weights of the final layer separate for each task.
We optimize the networks as in the permuted MNIST experiment and compare to five baseline networks with the same architecture trained on each task separately.

Overall, the online Laplace approximation in conjunction with a Kronecker factored approximation of the curvature achieves the highest test accuracy across all five tasks (see \aref{app:numerical:vision} for the numerical results).
However, the difference between the three Laplace-based methods is small in comparison to the improvement stemming from the better approximation to the Hessian.
We therefore plot the test accuracy curves through training only for the online Laplace approximation in the main text in \fref{fig:lenet} to show the difference to SI and between the two curvature approximations.
The corresponding figures for having a separate quadratic penalty for each task and the approximate Laplace approximation are in \aref{app:figures:vision}.

Using a diagonal Hessian approximation for the Laplace approximation, the network mostly remembers the first three tasks, but has difficulties learning the fifth one.
SI, in contrast, shows decaying performance on the initial tasks, but learns the fifth task almost as well as our method with a Kronecker factored approximation of the Hessian.
However, using the Kronecker factored approximation, the network achieves good performance relative to the individual networks across all five tasks.
In particular, it remembers the easier early tasks almost perfectly while being sufficiently flexible to learn the more difficult later tasks better than the diagonal methods, which suffer from forgetting.

\section{Conclusion} \label{sec:conclusion}

We proposed the online Laplace approximation, a Bayesian online learning method for overcoming catastrophic forgetting in neural networks.
By formulating a principled approximation to the posterior, we were able to substantially improve over EWC \citep{ewc} and SI \citep{synaptic:intelligence}, two recent methods that also add a quadratic regularizer to the objective for new tasks.
By further taking interactions between the parameters into account, we achieved considerable increases in test accuracy on the problems that we investigated, in particular for long sequences of datasets.
Our results demonstrate the importance of going beyond diagonal approximation methods which only measure the sensitivity of individual parameters.
Dealing with the complex interaction and correlation between parameters is necessary in moving towards a more complete response to the challenge of continual learning.

\bibliography{references}
\bibliographystyle{nips2018}

\onecolumn

\appendix

\section{Derivation of the Kronecker factorization of the diagonal blocks of the Hessian} \label{app:kf}

\citet{kfac} and \citet{kfra} both develop block-diagonal Kronecker factored approximations to the Fisher and Gauss-Newton matrix of fully connected neural networks respectively, which in turn both are positive semi-definite approximations of the Hessian.
Both use their approximations for optimization, hence the positive semi-definiteness is crucial in order to prevent parameter updates that increase the loss.
We require this property as well, as we perform a Laplace approximation and the Normal distribution requires its covariance to be positive semi-definite.

In the following, we provide the basic derivation for the diagonal blocks of the Hessian being Kronecker factored as developed in \citep{kfra} and state the recursion for calculating the pre-activation Hessian.

We denote a neural network as taking an input $\act_0 = x$ and producing an output $\preact_L$.
The input is passed through layers $l=1,\ldots,L$ as linear pre-activations $\preact_l = \weight_l \act_{l-1}$ and non-linear activations $\act_l = f_l(\preact_l)$, where $\weight_l$ denotes the weight matrix and $f_l$ the elementwise activation function.
Bias terms can be absorbed into $\weight_l$ by appending a $1$ to every $\act_l$.
The weights are optimized w.r.t. an error function $\err(y, \preact_L)$, which can usually be expressed as a negative log likelihood.

Using the chain rule, the gradient of the error function w.r.t. an individual weight can be calculated as:

\begin{equation}
    \ederiv{\weight_{a,b}^l}
    = \sum_i \deriv{\preact^l_i}{\weight^l_{a,b}} \ederiv{\preact^l_i}
    = \act^{l-1}_b \ederiv{\preact^l_a}
\end{equation}

Differentiating again w.r.t. another weight within the same layer gives:

\begin{equation}
    \left[ \hess_l \right]_{(a,b), (c,d)}
    \equiv \ehess{\weight_{a,b}}{\weight_{c,d}}
    = \act^{l-1}_b \act^{l-1}_d \left[ \xhess_l \right]_{(a,c)}
\end{equation}

where

\begin{equation}
    \left[ \xhess_l \right]_{a,b} \equiv \ehess{\preact^l_a}{\preact^l_b} 
\end{equation}

is defined to be the pre-activation Hessian.

This can also be expressed in matrix notation as a Kronecker product:

\begin{equation}
    \hess_l
    = \ehess{\vect(\weight^l)}{\vect(\weight^l)}
    = \left( \act_{l-1} \act_{l-1}^\top \right) \otimes \xhess_l
\end{equation}

Similar to backpropagation, the pre-activation Hessian can be calculated as:

\begin{equation}
    \xhess_l = B_l \weight_{l+1}^\top \xhess_{l+1} \weight_{l+1} B_l + D_l
\end{equation}

where the diagonal matrices $B_l$ and $D_l$ are defined as

\begin{align}
    B_l &= \diag(f_l'(\preact_l))\\
    D_l &= \diag(f_l''(\preact_l) \ederiv{\act_l})
\end{align}

$f'$ and $f''$ denote the first and second derivative of $f$.
The recursion for $\xhess$ is initialized with the Hessian of the error w.r.t. the network outputs, i.e. $\xhess_L \equiv \ehess{h_L}{h_L}$.
For the derivation of the recursion and how to calculate the diagonal blocks of the Gauss-Newton matrix, we refer the reader to \citep{kfra}, and to \citep{kfac} for the Fisher matrix.

\section{Visualization of the effect of $\lambda$ for a Gaussian prior and posterior} \label{app:reg:vis}

\begin{figure}[ht!]
    \centering
    \includegraphics[width=0.5\linewidth]{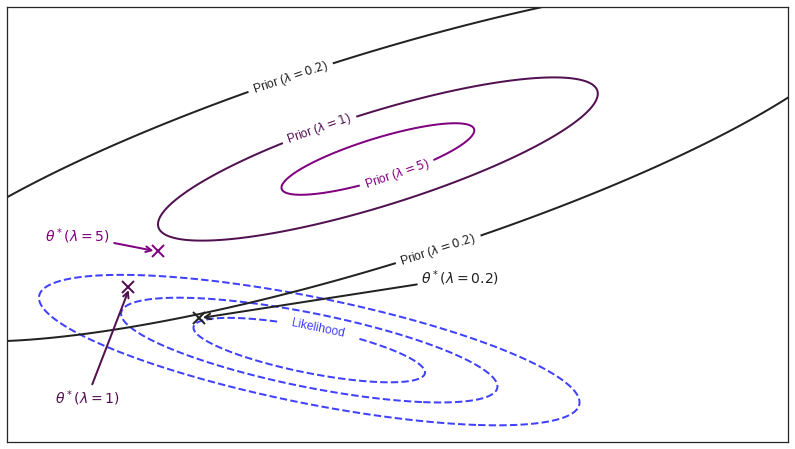}
    \caption{Contours of a Gaussian likelihood (dashed blue) and prior (shades of purple) for different values of $\lambda$. Values smaller than $1$ shift the joint maximum $\theta^*$, marked by a `${\times}$',towards that of the likelihood, i.e. the new task, for values greater than $1$ it moves towards the prior, i.e. previous tasks.}
    \label{fig:contours}
\end{figure}

A small $\lambda$ resulting in high uncertainty shifts the mode towards that of the likelihood, i.e. enables the network to learn the new task well even if our posterior approximation underestimates the uncertainty.
Vice versa, increasing $\lambda$ moves the joint mode towards the prior mode, improving how well the previous parameters are remembered.
The optimal choice depends on the true posterior and how closely it is approximated.

In principle, it would be possible to use a different value $\lambda_t$ for every dataset.
In our experiments, we keep the value of $\lambda$ the same across all tasks as the family of posterior approximation is the same throughout training.
Furthermore, using a separate hyperparameter for each task would let the number of hyperparameters grow linearly in the number of tasks, which would make tuning them costly.

\section{Additional related work} \label{app:related}

Various methods for overcoming catastrophic forgetting without a Bayesian motivation have also been proposed over the past year.
\citet{synaptic:intelligence} develop `Synaptic Intelligence' (SI), another quadratic penalty on deviations from previous parameter values where the importance of each weight is heuristically measured as the path length of the updates on the previous task.
\citet{gem} formulate a quadratic program to project the gradients such that the gradients on previous tasks do not point in a direction that decreases performance; however, this requires keeping some previous data in memory.
\citet{deep:generative:replay} suggest a dual architecture including a generative model that acts as a memory for data observed in previous tasks.
Other approaches that tackle the problem at the level of the model architecture include \citep{progessive:nns}, which augments the model for every new task, and \citep{pathnet}, which trains randomly selected paths through a network.
\citet{hard:attention} propose sharing a set of weights and modifying them in a learnable manner for each task.
\citet{conceptor} introduce conceptor-aided backpropagation to shield gradients against reducing performance on previous tasks.

\section{Optimization details} \label{app:details}

For the permuted MNIST experiment, we found the performance of the methods that we compared to mildly depend on the choice of optimizer.
Therefore, we optimize all techniques with Adam \citep{adam} for $20$ epochs per dataset and a learning rate of $10^{-3}$ as in \citep{synaptic:intelligence}, SGD with momentum \citep{momentum} with an initial learning rate of $10^{-2}$ and $0.95$ momentum, and Nesterov momentum \citep{nag} with an initial learning rate of $0.1$, which we divide by $10$ every $5$ epochs, and $0.9$ momentum.
For the momentum based methods, we train for at least $10$ epochs and early-stop once the validation error does not improve for $5$ epochs.
Furthermore, we decay the initial learning rate with a factor of $\frac{1}{1 + k t}$ for the momentum-based optimizers, where $t$ is the index of the task and $k$ a decay constant.
We set $k$ using a coarse grid search for each value of the hyperparameter $\lambda$ in order to prevent the objective from diverging towards the end of training, in particular with the Kronecker factored curvature approximation.
For the Laplace approximation based methods, we consider $\lambda \in \{1, 3, 10, 30, 100\}$; for SI we try $c \in \{0.01, 0.03, 0.1, 0.3, 1\}$.
We ultimately pick the combination of optimizer, hyperparameter and decay rate that gives the best validation error across all tasks at the end of training.
For the Laplace-based methods, we found momentum based optimizers to lead to better performance, whereas Adam gave better results for SI.

\section{Numerical results of the vision experiment} \label{app:numerical:vision}

\begin{table}[h]
    \centering
    \caption{Per dataset test accuracy at the end of training on the suite of vision datasets. SI is Synaptic Intelligence \citep{synaptic:intelligence} and EWC Elastic Weight Consolidation \citep{ewc}. We abbreviate Per-Task Laplace (one penalty per task) as PTL, Approximate Laplace (Laplace approximation of the full posterior at the mode of the approximate objective) and our Online Laplace approximation as OL. nMNIST refers to notMNIST, fMNIST to FashionMNIST and C10 to CIFAR10.}
    \begin{tabular}{ll|ccccc|r}
         && \multicolumn{6}{c}{Test Error (\%)} \\
         Method & Approximation & MNIST & nMNIST & fMNIST & SVHN & C10 & Avg.\\
         \hline
         SI & n/a & 87.27 & 79.12 & 84.61 & 77.44 & 57.61 & 77.21\\
         PTL & Diagonal (EWC) & 97.83 & 94.73 & 89.13 & 79.80 & 53.29 & 82.96\\
         & Kronecker factored & 97.85 & 94.92 & 89.31 & 85.75 & 58.78 & 85.32\\
         AL & Diagonal & 96.56 & 92.33 & 89.27 & 78.00 & 56.57 & 82.55\\
         & Kronecker factored & 97.90 & 94.88 & 90.08 & 85.24 & 58.63 & 85.35\\
         OL & Diagonal & 96.48 & 93.41 & 88.09 & 81.79 & 53.80 & 82.71\\
         & Kronecker factored & 97.17 & 94.78 & 90.36 & 85.59 & 59.11 & 85.40
    \end{tabular}
    \label{tab:vision:test}
\end{table}

\vfill

\section{Additional figures for the vision experiment} \label{app:figures:vision}

\begin{figure}[!ht]
    \centering
    \begin{subfigure}[]{\textwidth}
        \includegraphics[width=\textwidth]{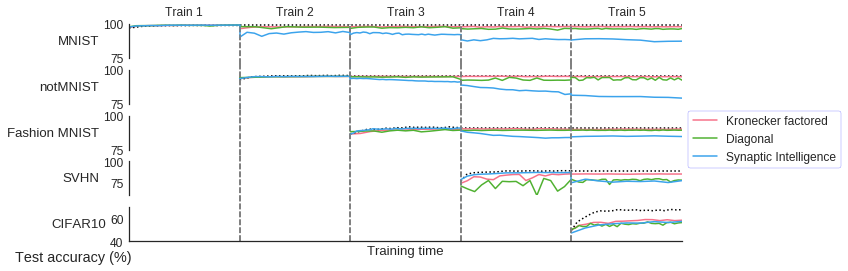}
        \caption{Approximate Laplace}
    \end{subfigure}%
    \\%
    \begin{subfigure}[]{\textwidth}
        \includegraphics[width=\textwidth]{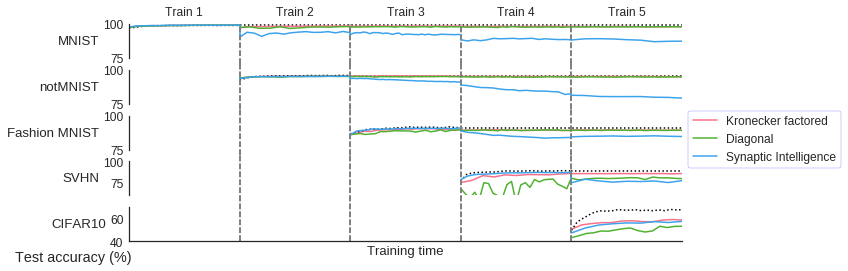}
        \caption{Per-task Laplace}
    \end{subfigure}
    \caption{Test accuracy of a convolutional network on a sequence of vision datasets for different methods for preventing catastrophic forgetting. We train on the datasets separately in the order displayed from top to bottom and show the network's accuracy on each dataset once training on it has started. The dotted black line indicates the performance of a network with the same architecture trained separately on the task.}
\end{figure}

\end{document}